# Video Forgery Detection for Surveillance Cameras: A Review

Noor B. Tayfor [1], Tarik A. Rashid [2, 4 *], Shko M. Qader [3], Hussein M. Ali [4 *], Abdulhady A. Abdullah [4], Bryar A. Hassan [2], Mohammed H. Abdalla [5], Jafar Majidpour [5], Aram M. Ahmed [2], Aso M. Aladdin [6], Ahmed S. Shamsaldin [2], Azad A. Ameen [6], Mahmood Yashar Hamza [7], and Janmenjoy Nayak [8]

[1]Department of Information Technology, Kurdistan Technical Institute, Sulaimani, Iraq. noorbahjattay-for@gmail.com

[2]Computer Science and Engineering Department, University of Kurdistan Hewlêr, Erbil, Iraq. ( tarik.ahmed@ukh.edu.krd, bryar.ahmad@ukh.edu.krd, aram.mahmood@ukh.edu.krd, ah-med.saadaldin@ukh.edu.krd )

[3]Department of Computer Science, Kurdistan Technical Institute, Sulaimani, Iraq. shkomq@gmail.com

[4]Artificial Intelligence and Innovation Centre, University of Kurdistan Hewler, Erbil, Iraq. ( hussein.mohammedali@ukh.edu.krd, abdulhady.abas@ukh.edu.krd )

[5]Department of Computer Science, University of Raparin, Rania, Iraq. ( mohammed.meera@uor.edu.krd, jafar.majidpoor@uor.edu.krd )

[6]Department of Software Engineering, College of Engineering and Computational Science, Charmo University, Chamchamal 46023, Iraq.

( aso.aladdin@charmouniversity.org, azad.ameen@chu.edu.iq )

[7]Department of Computer Engineering, Tishk International University, Erbil, Iraq. mahmood.yashar@tiu.edu.iq

[8]Department of Computer Science, Maharaja Sriram Chandra Bhanja Deo University, Baripada, Odisha, India. mailforjnayak@gmail.com

*Correspondence: tarik.ahmed@ukh.edu.krd. Tel.: +964 (0) 750 109 4233, 44011.

*Correspondence: hussein.mohammedali@ukh.edu.krd. Tel.: +964 (0) 751 126 6540, 44011.

## Abstract

The widespread availability of video recording through smartphones and digital devices has made video-based evidence more accessible than ever. Surveillance footage plays a crucial role in security, law enforcement, and judicial processes. However, with the rise of advanced video editing tools, tampering with digital recordings has become increasingly easy, raising concerns about their authenticity. Ensuring the integrity of surveillance videos is essential, as manipulated footage can lead to misinformation and undermine judicial decisions. This paper provides a comprehensive review of existing forensic techniques used to detect video forgery, focusing on their effectiveness in verifying the authenticity of surveillance recordings. Various methods, including compression-based analysis, frame duplication detection, and machine learning-based approaches, are explored. The findings highlight the growing necessity for more robust forensic techniques to counteract evolving forgery methods. Strengthening video forensic capabilities will ensure that surveillance recordings remain credible and admissible as legal evidence.

## 1. Introduction

The presence of considerably low-priced digital video cameras on smart devices and gadgets and the availability of many websites, play a remarkably vital role in sharing various videos. YouTube, Facebook, Twitter, Instagram, TikTok, Snapchat, WhatsApp, etc. [1-2] are used daily to share huge amounts of recorded videos. In today's technologically advanced world, digital video recorders, particularly security cameras, are widely accessible and create a vast amount of multimedia footage [3]. Surveillance systems are routinely used to ensure public safety. The security camera has achieved massive popularity as an efficient safety precaution in offices, residences, and numerous public locations. Although video manipulating software, for example, Adobe Premier, Adobe After Effect, PowerDirector Essential, GNU Gimp, Blender, Vegas, etc., allows users to create a fabricated video quickly. Therefore, the integrity of surveillance videos cannot be guaranteed because they are no longer regarded as crucial legal evidence, mainly when that video footage is treated as evidence against any crimes in court. Moreover, it has been verified that forensic analysis of videos must be completed to prove the authenticity of their contents [4]. As a result, authenticating surveillance recordings and identifying the forging process is critical [5]. Video forgery refers to any intentional change of digital video for fabrication. At the same time, Digital video forgery detection refers to determining if digital video footage has been deliberately changed. Thus, this survey article targets to review the recent state-of-the-art methods and approaches introduced to detect video tampering in surveillance videos and highlight their limitations since no prior work has reviewed the methods and techniques utilized to detect a forgery in videos obtained from surveillance cameras. As depicted in Figure 1, the number of publications published to detect tampering in surveillance videos was not significantly great compared to the publications that frequently deal with the issues of detecting forgery in videos. In the years 2017 and 2022, three articles were published. However, just two research have been done in 2013, 2019, 2020, and 2021. While only one article was published in 2014 and 2018. Surprisingly, no work has been published in 2015 and 2016 related to forgery detection in surveillance videos.

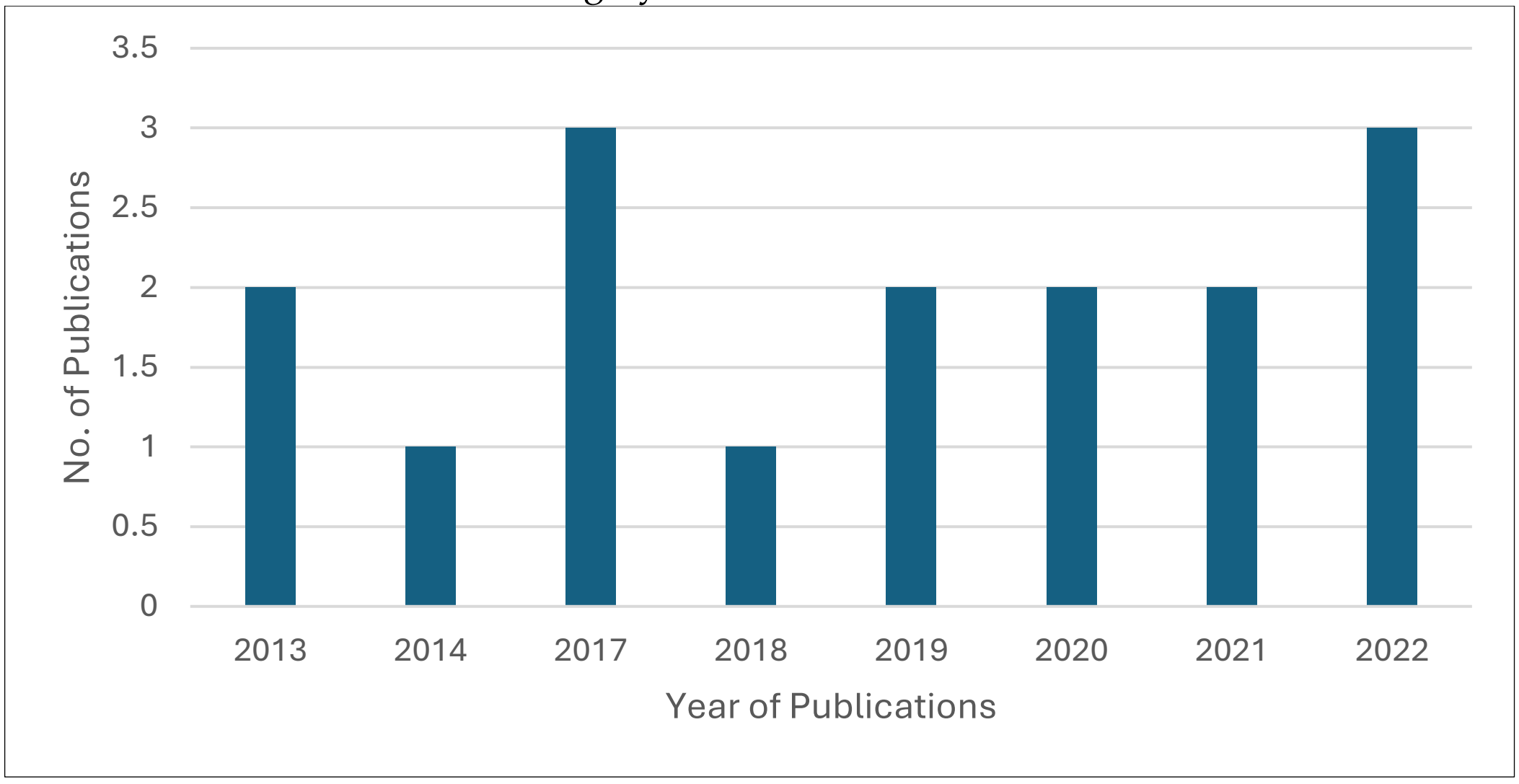

**Figure 1.** Summary of Publications for Forgery Detection in Surveillance Video since 2013.

This paper is organized as follows: section 3 includes the approaches that are frequently used to detect a forgery in videos; Section 4 determines the popular types of video forgery; Section 5 states the techniques and methods of detecting tampering in surveillance videos; Section 6 discusses the best mechanism that has been proposed to detect fraud in footage videos; and finally, Section 7 includes the conclusion and future work.

## 2. The Mechanisms of Video Forgery Detection

Active and passive strategies (i.e., represented in Figure 2) are two kinds of forgery detection systems in digital video:

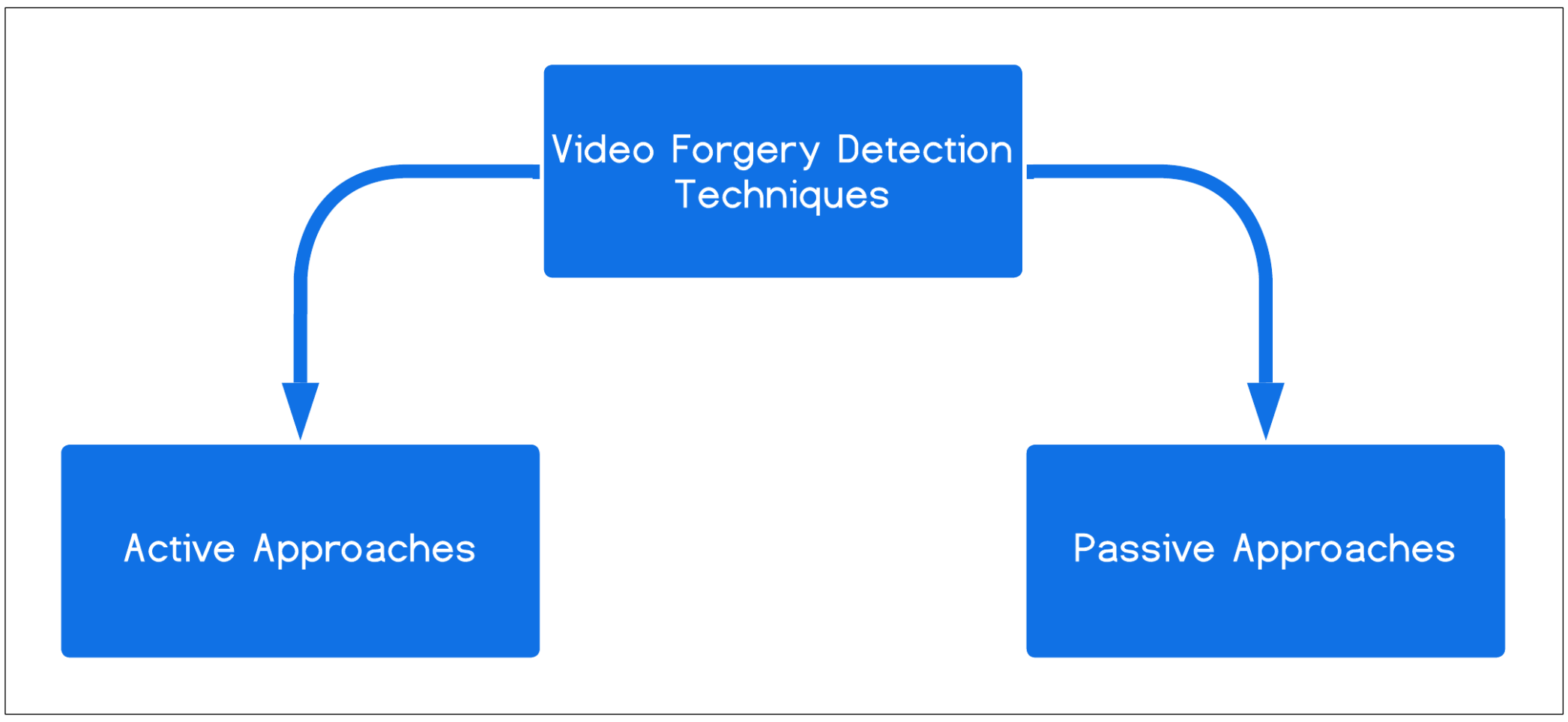

**Figure 2.** Video Forgery Detection Approaches.

*2.1. Active Approaches*

Active techniques for detecting forgeries employ pre-embedded information, for example, a digital watermark [6] or digital signature [7], to assess the authenticity of content ownership, integrity, and copyright violations [8], as shown clearly in Figure 3.

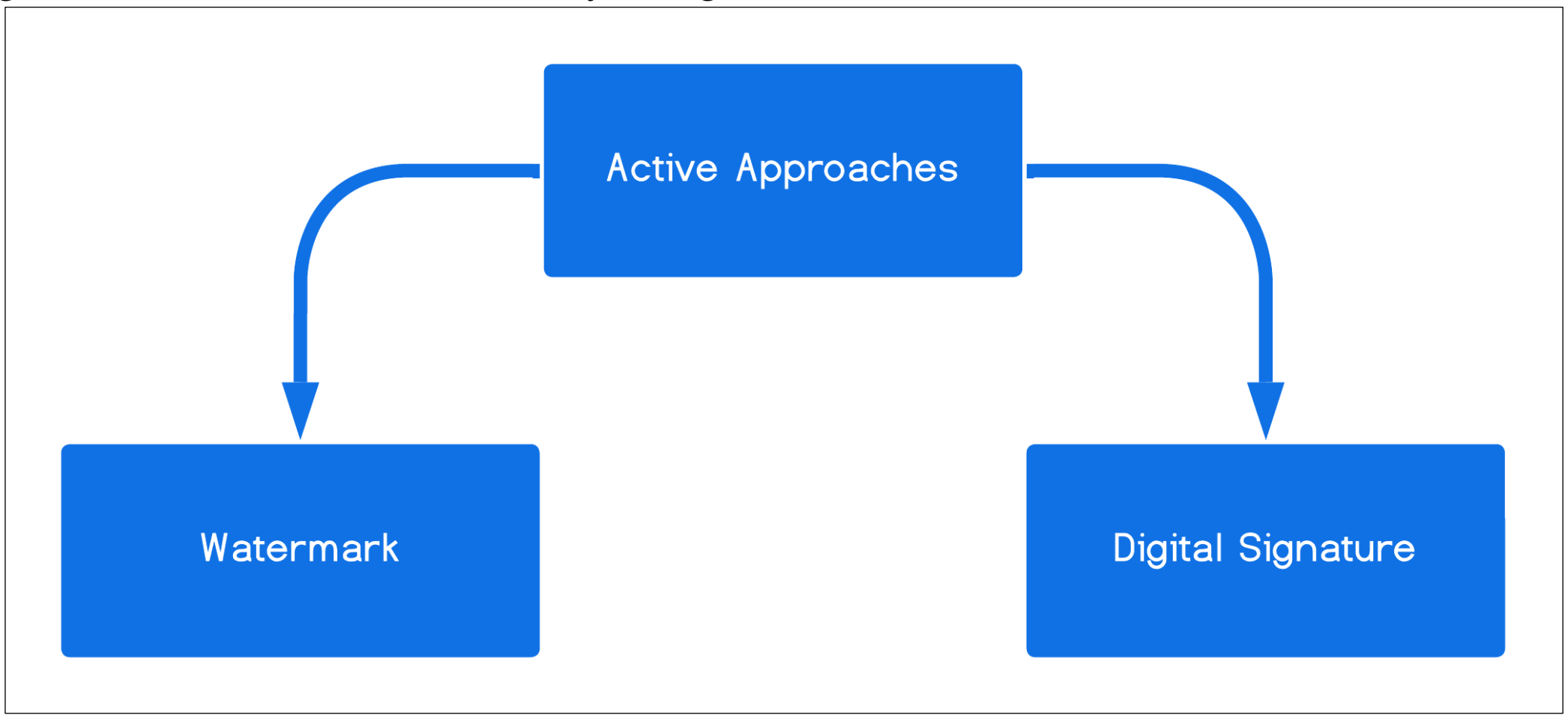

**Figure 3.** Active Techniques.

If a video's content has been altered, the encoded watermark or signature changes, indicating that the video has been tampered with [9]. The disadvantages of active methods are the need for particular hardware, such as a camera, and the requirement of embedding a digital signature or digital watermark during the recording stage, as illustrated in Figures 4 and 5.

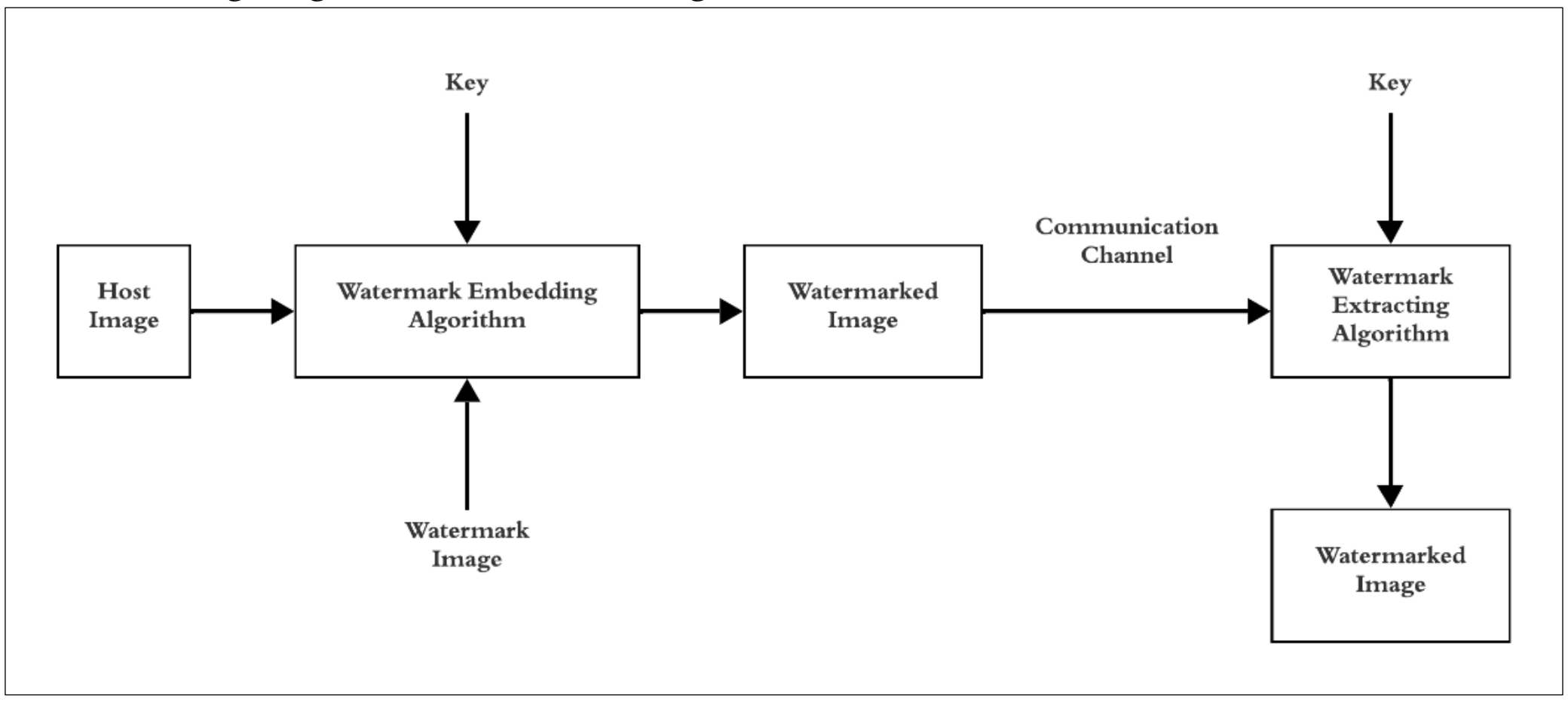

**Figure 4.** Watermarking Process.

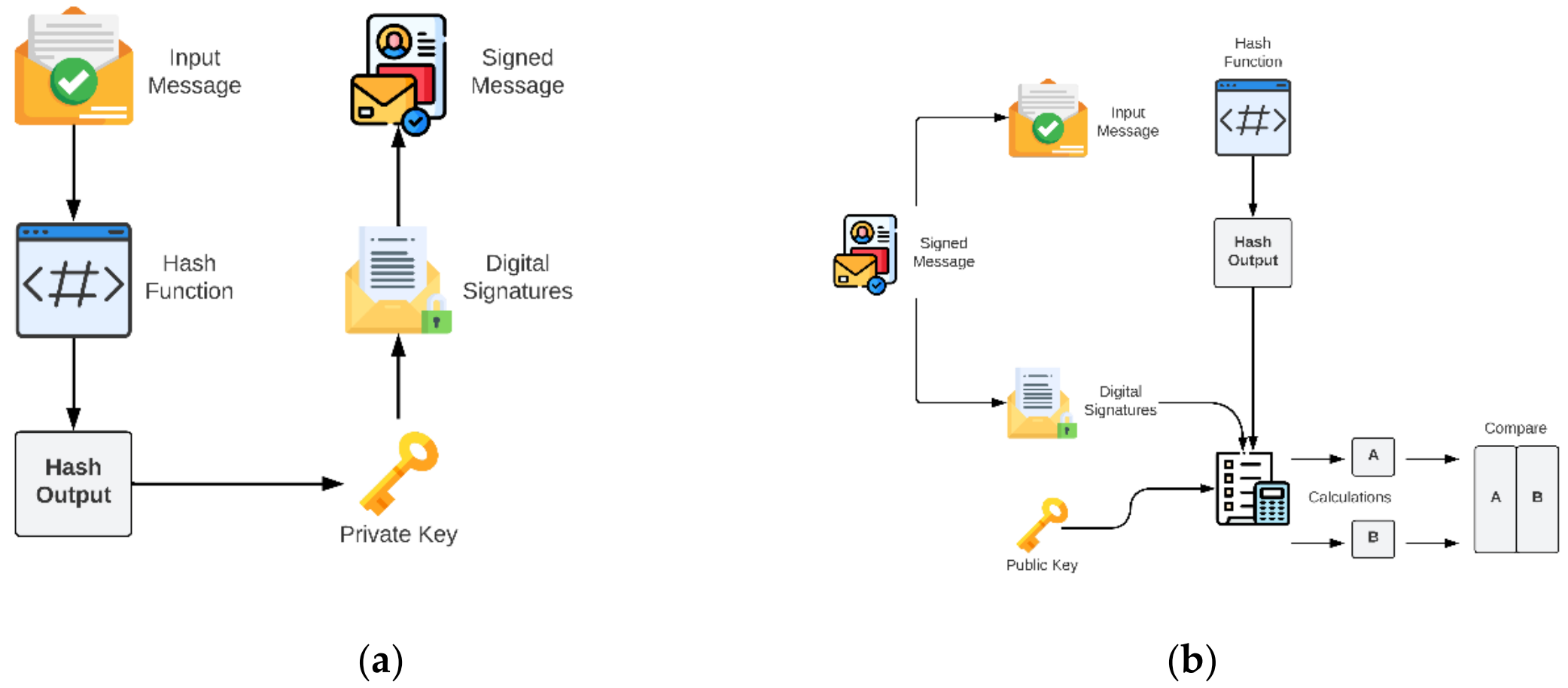

**Figure 5.** Digital Signature Process: (**a**) Signing of the Message. (**b**) Verification of the Message.

*2.2. Passove Approaches*

Passive forgery detection methods are a promising digital security path forward [12]. During the absence of pre-embedded data, passive forgery detection depends on the inherent properties of the digital video rather than information that could be utilized to confirm the authenticity of the video [13]. However, since the majority of videos lack pre-embedded information, for instance, a signature or watermark, it is difficult to detect modifications using an active technique. Because this method does not require a specific tool and prior knowledge of the video's content and generates static and temporal distortions in a video that must be verified to detect modified videos. Consequently, it is also referred to as the Passive-Blind Technique [14]. Passive methods (i.e., represented in Figure 6), include the following sorts of altering parts based on the video's regional properties:

1. Spatial Tampering
2. Temporal Tampering
3. Spatio-Temporal Tampering
4. Re-Projection

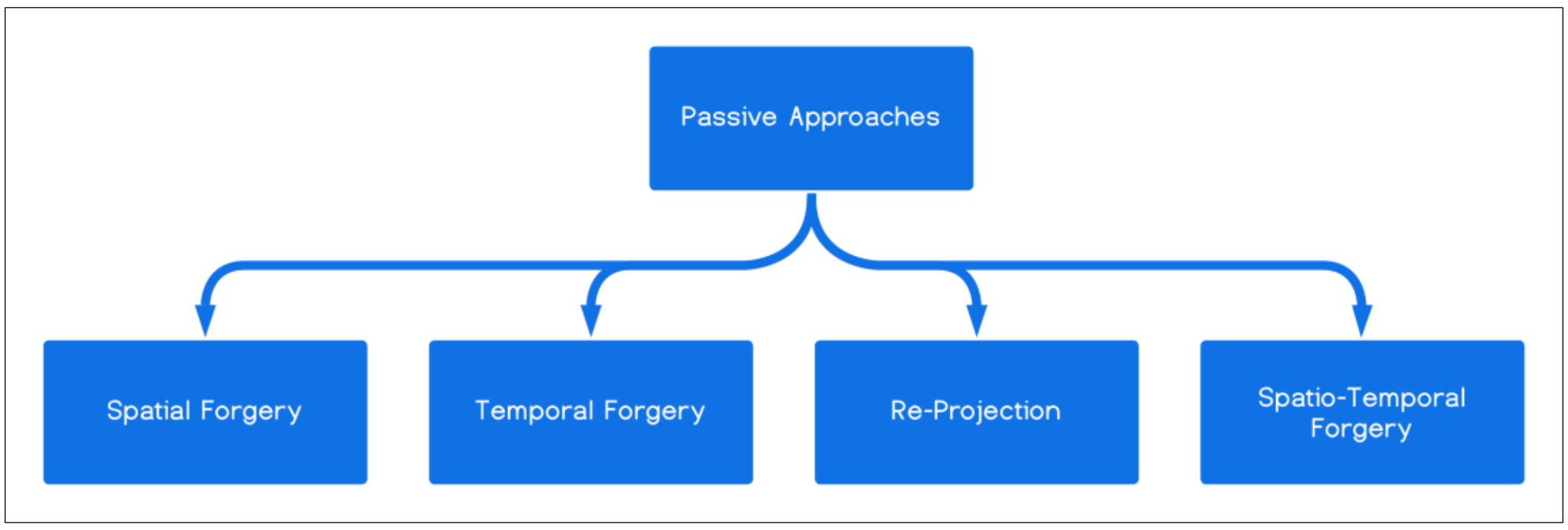

**Figure 6.** Passive Techniques.

2.2.1. Spatial Tampering/ Intra-frame Forgery

Spatial Tampering or Intra-frame forgery determines the kind of counterfeit that involves manipulating the original contents of specific frames [15]. It can be accomplished by modifying the pixel bits in a frame or the adjacent ones in a video sequence (i.e., along the x-y axis) [16]. Figure 7 shows how the spatial approach could be applied to obtain a fake video. *

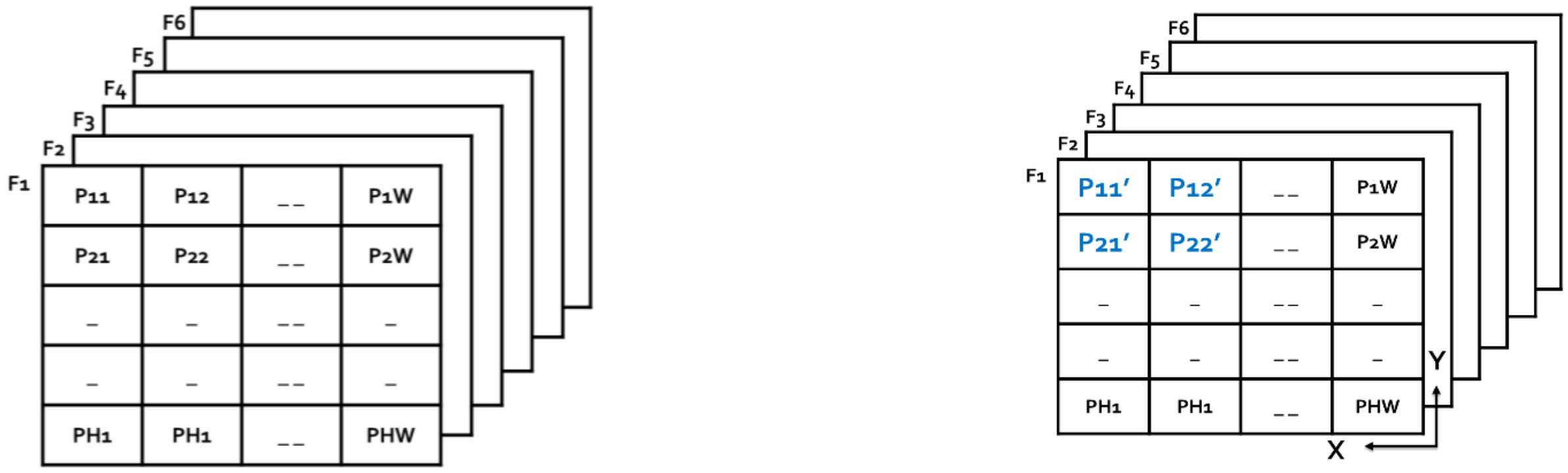

(a) (b)

**Figure 7.** Spatial Video Tampering: (**a**) Original Video. (**b**) Spatial Video Tampering.

Forensic professionals have access to many sorts of information (artifacts or footprints) to detect spatial tampering and localization. These details reveal, represented in Table 1, that the techniques fall under the following groups of techniques: deep learning techniques [18]–[23], camera source features techniques [24]–[27], pixels and source device [28]–[30], SVD techniques [31], compression techniques [32]–[34] and statistical features techniques [35]–[37]. It could be classified (i.e., as shown in Figure 8) into:

1. Copy-Move Forgery
2. Splicing Forgery
3. Upscale Crop Forgery

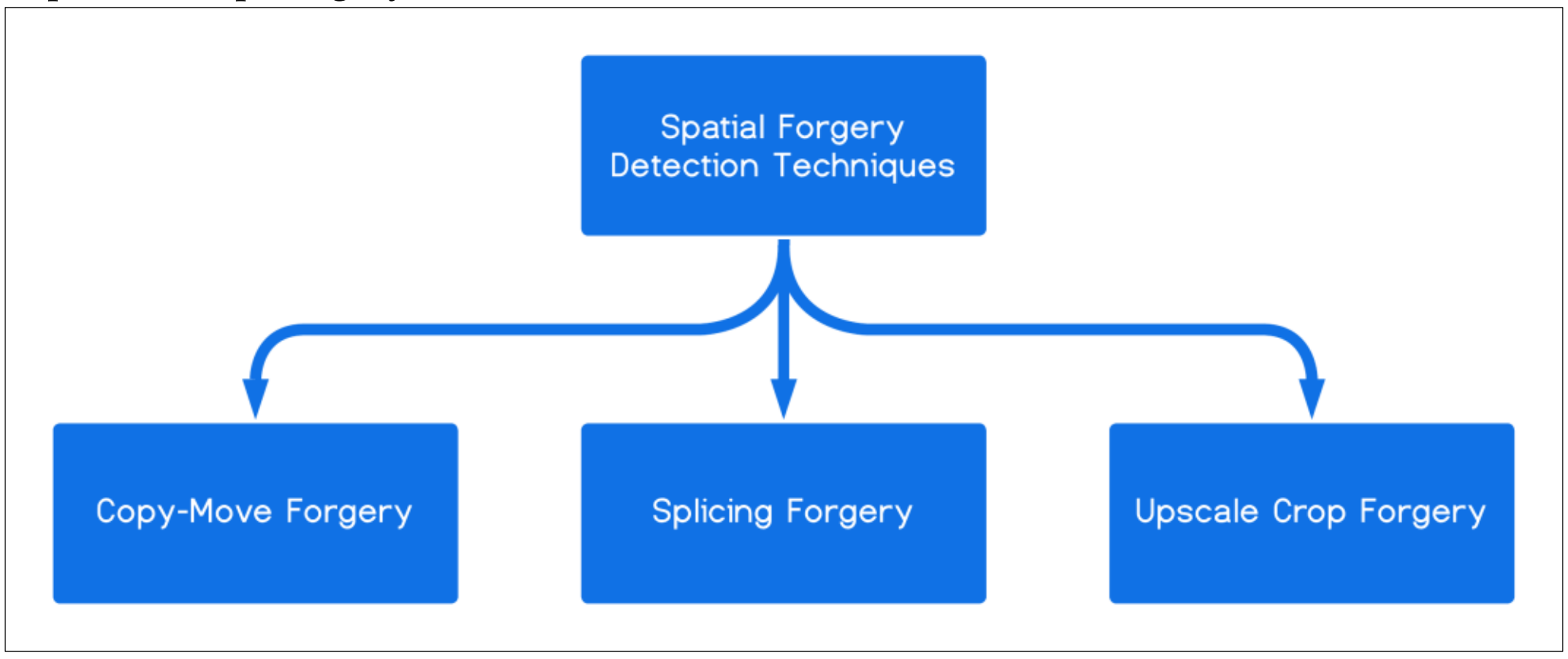

**Figure 8.** Spatial (Intra-Frame) Tampering Detection Methods.

The groups of Spatial (Intra-Frame) forgery detection methods are listed below:

1. Deep Learning Techniques
2. Camera Source Features Techniques
3. Pixels And Source Device Techniques
4. SVD Techniques
5. Compression Techniques
6. Statistical Features Techniques

Copy-Move forgery is one of the most widespread forms of digital image/video tampering [38]. An object can be inserted or deleted from a video scene using this sort of forgeries. Simultaneously, It may be used to replicate video elements by copying a section of the video frame and pasting it elsewhere, either within the same or a separate video frame. [39]. This procedure can popularly be used to hide the desired location in the frame [40], [41]. As a result, it is also known as copypaste or area manipulation forgery. Figure 9 depicts a. the original image that has been taken and b. the final image after applying copy move tampering where the tree has been copied next to the tree shown on the left-hand side.

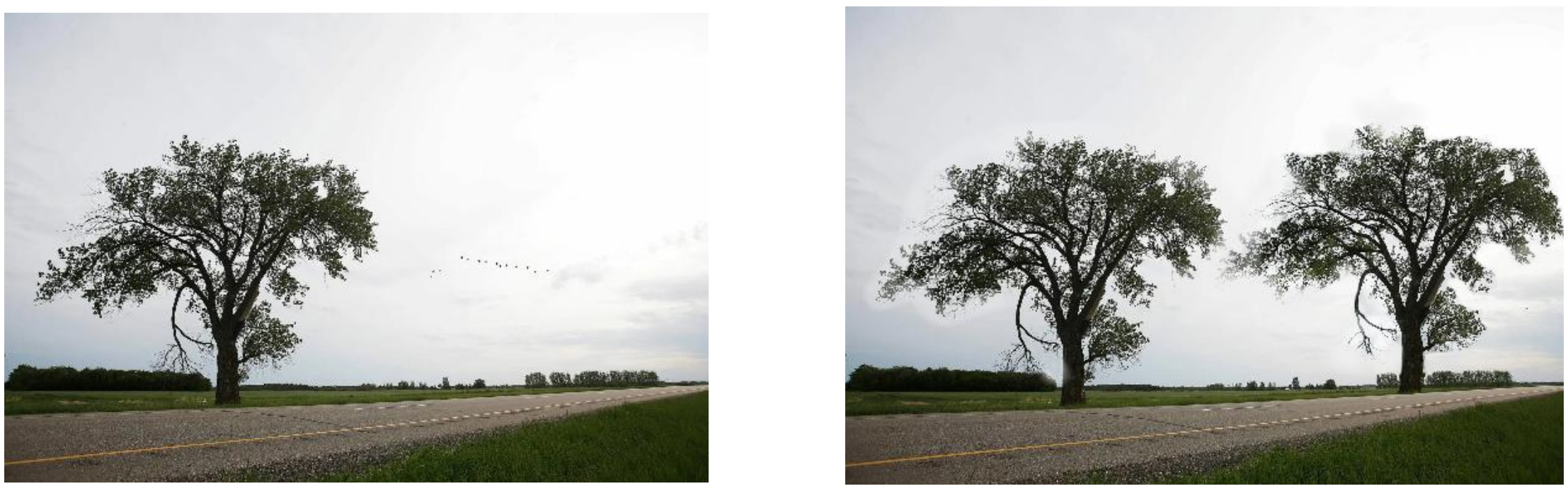

(**a**) (**b**)

**Figure 9.** Copy Move Forgery Example: (**a**) Original Video. (**b**) Copy Move Forgery.

Splicing forgery is another type of forgery, where the new frame is created by copying and pasting a fragment of an existing video frame [43-46]. As illustrated in Figure 10, The new spliced video has been composed by merging the two video frames. This forgery is often hard to identify as the generated video is compressed, resampled, and blurred. Besides, it is used for malicious purposes. Thus, developing trustworthy splicing detection tools to assess the veracity of photos has emerged as a key challenge. This inspires the researchers to develop several methods to identify splicing forgeries. The main objective of many ways to identify image splicing is to detect the region of abnormalities using characteristics of the image [47,48].

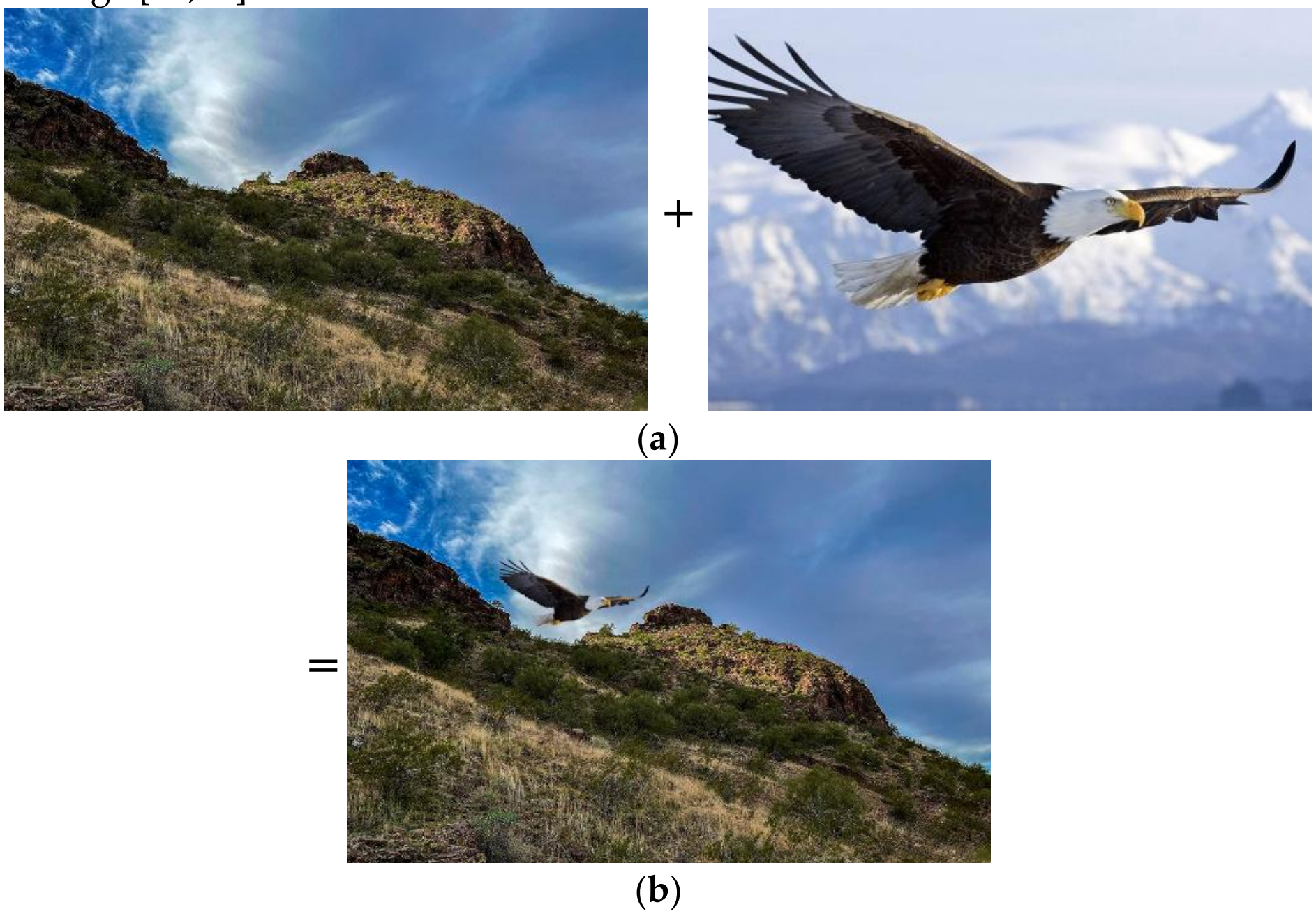

**Figure 10.** Explanation of Splicing Forgery: (**a**) Original Video. (**b**) Spliced Video.

On the other hand, Upscale crop forgery refers to cropping and enlarging the outer portion of a video frame to remove an area or object that might represent any evidence of incriminating events [50] [51]. Figure 11 depicts an example of an upscale crop video tampering where Figure 11 (a) shows the original scene whilst Figure 11 (b) illustrates the fake video created in Figure 11 (a) where the walking lady has been removed.

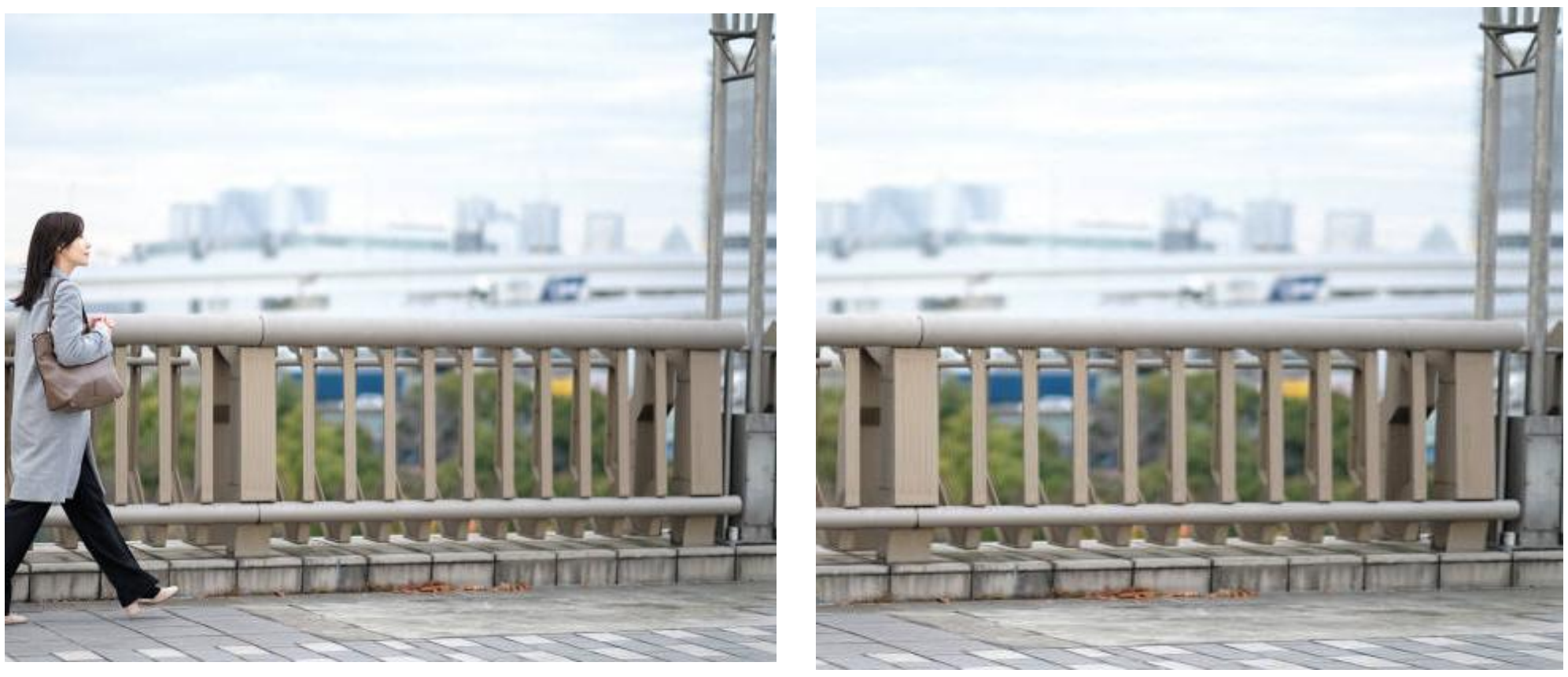

(**a**) (**b**)

**Figure 11.** Upscale Crop Video Forgery: (**a**) Original Video. (**b**) Altered Video after applying Upscale Crop Forgery.

2.2.2. Temporal Tampering/ Inter-frame Forgery

Temporal Tampering or Inter-frame forgery manipulation is done on the video's concatenated chain of frame sequence either through the involvement of replacement, reordering, addition, or removal of a video frame [53]. The actions that can be included in this tampering are generally executed at the frame level [54]. Figure 12 represents the process of creating temporal video tampering.

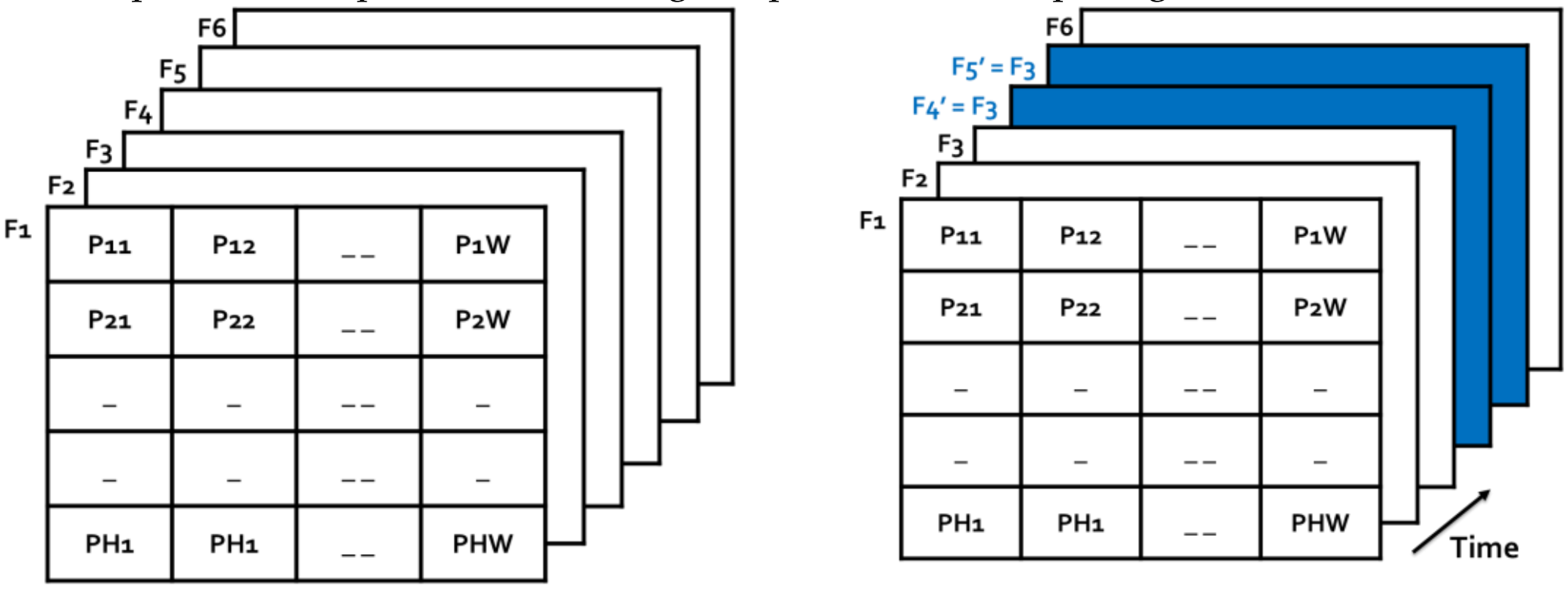

(**a**) (**b**)

**Figure 12.** Temporal Video Tampering: (**a**) Original Video. (**b**) Temporal Video Tampering.

The following groups can be used to categorise the algorithms used to identify temporal forgery, depicted in Table 2, for instance, statistical features [55]–[58], frequency domain features [59]–[63], residual and optical flow [64]–[70], pixel and texture [71]–[76], and deep learning [77]–[82] strategies. The various types of temporal forgery are illustrated in Figure 13.

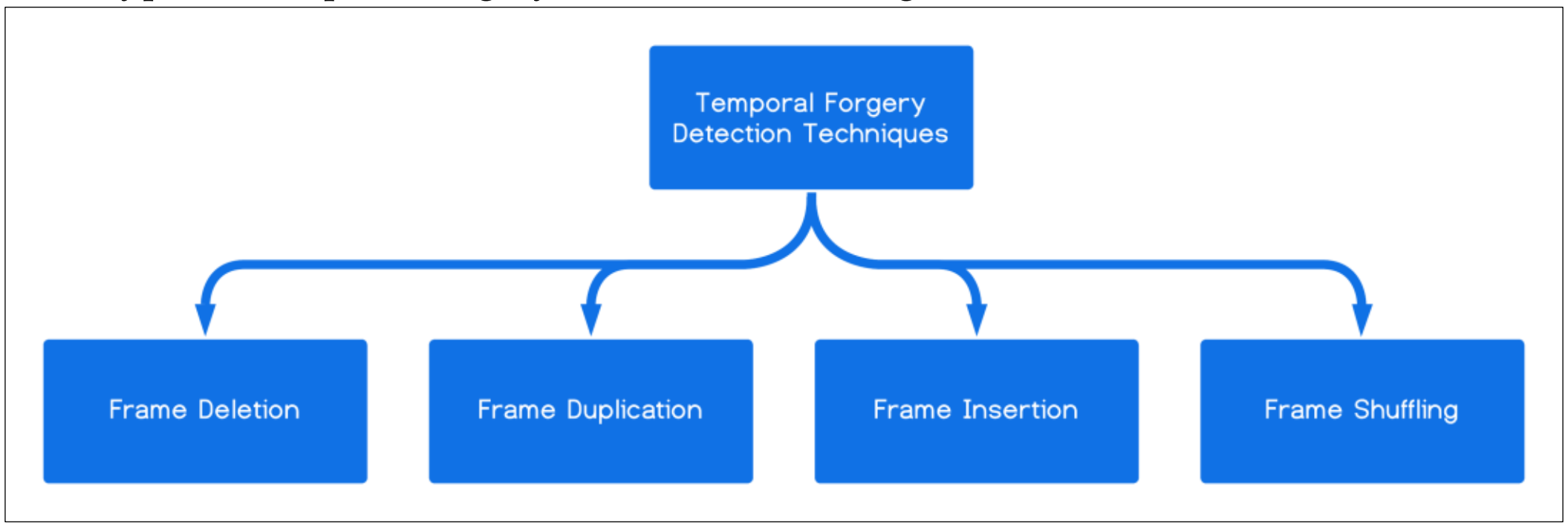

**Figure 13.** Temporal (Intra-Frame) Tampering Detection Methods.

The groups of Temporal (Inter-Frame) Forgery Detection Methods are listed below as well:

1. Statistical Features Techniques
2. Frequency Domain Features Techniques
3. Residual And Optical Flow Techniques

4. Pixel And Texture Features Techniques
5. Deep Learning Techniques

Frame deletion is a type of temporal tampering/ inter-frame forgery, which tends to of manipulation, this manipulation eliminates frames from a video on purpose to manufacture misleading proof of illegal activities [83]. Figure 14 illustrates the sequence of the forged video before and after involving the frame deletion tampering.

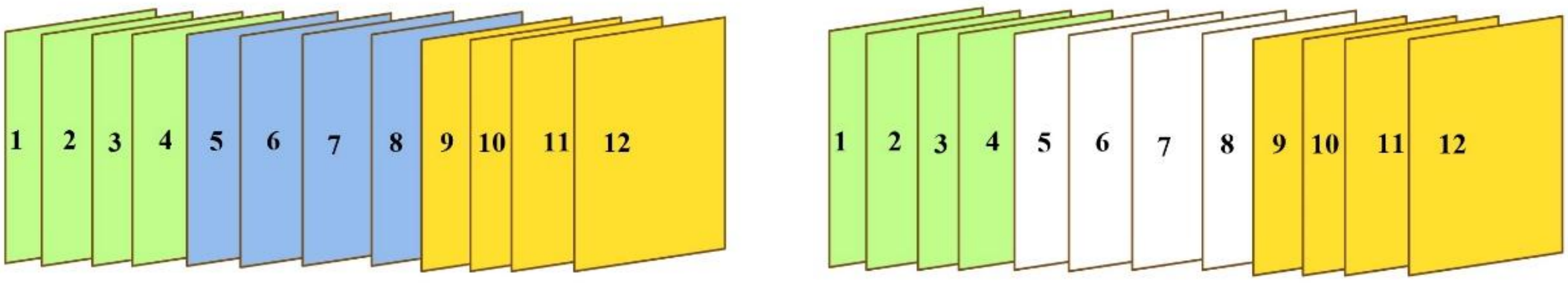

(**a**) (**b**)

**Figure 14.** Frame Deletion Forgery: (**a**) Original Video Sequence. (**b**) Frame Deletion Tampering.

Frame duplication is another type of tampering, this counterfeit deliberately repeats some of the frames in a movie [85]. Figure 15 clarifies the sequence of the forged video before and after involving the frame duplication tampering.

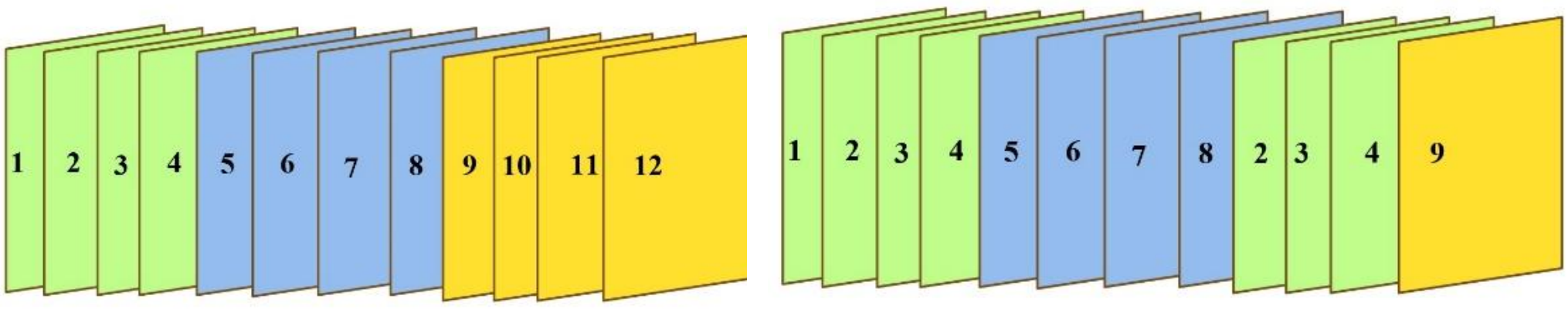

(**a**) (**b**)

**Figure 15.** Frame Duplication Forgery: (**a**) Original Video Sequence. (**b**) Frame Duplication Tampering.

There is another form of frame duplication forgery that is called frame mirroring [86] which aims to copy some frames from the original video and paste them randomly into other locations in the same video. Figure 16 delineates the sequence of the forged video after involving the frame mirroring tampering where (a) is the original video and (b) is the counterfeit video after frame mirroring in which the mirrored copies of the 2nd, 3rd, and 4th frames are passed between the frame 6th and 7th location.

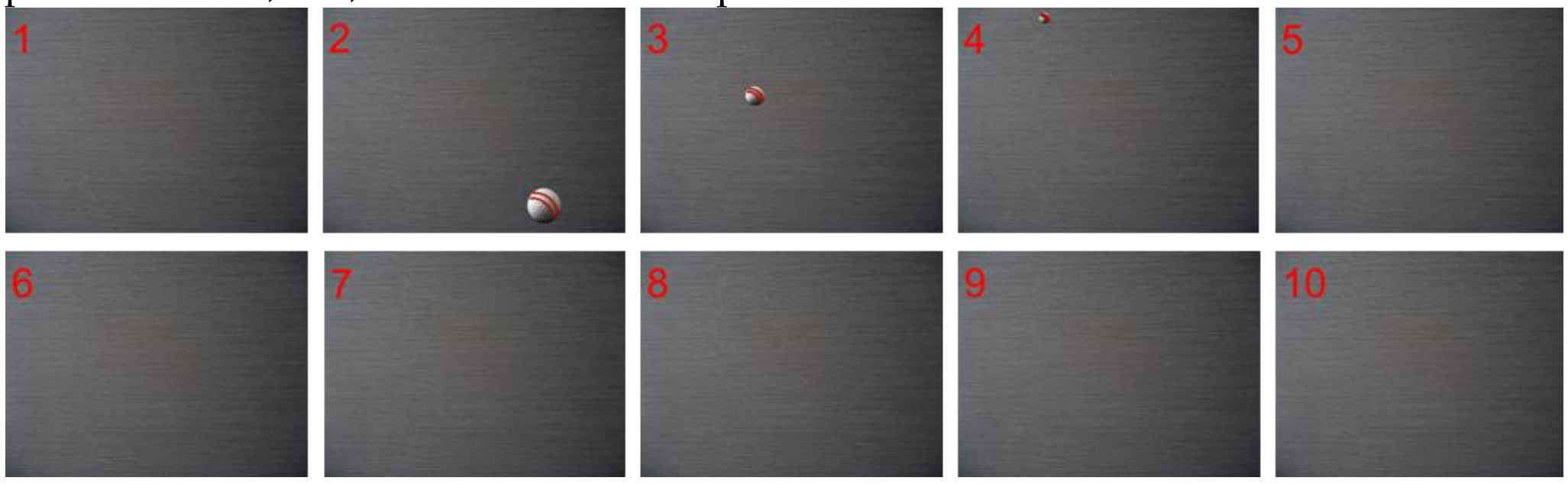

(**a**)

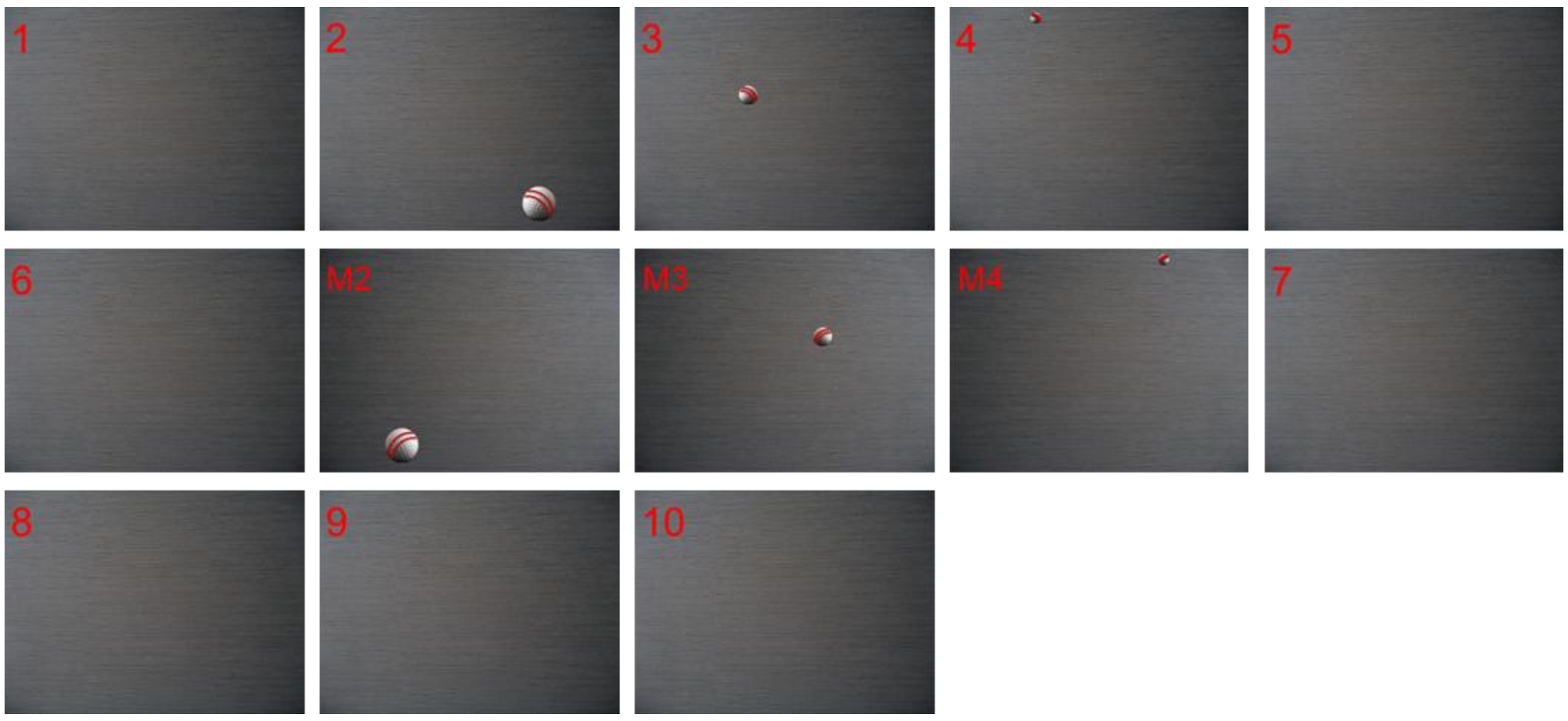

(**b**)

**Figure 16.** Frame Mirroring Tampering: (**a**) Original Video Sequence. (**b**) Frame Monitoring Tampering.

Also, there is frame insertion, which is used mainly for any criminal conduct or fraudulent proof, frames from other different videos or the same video are randomly placed at positions [87]. Figure 17 explains the sequence of the forged video before and after involving the frame insertion tampering.

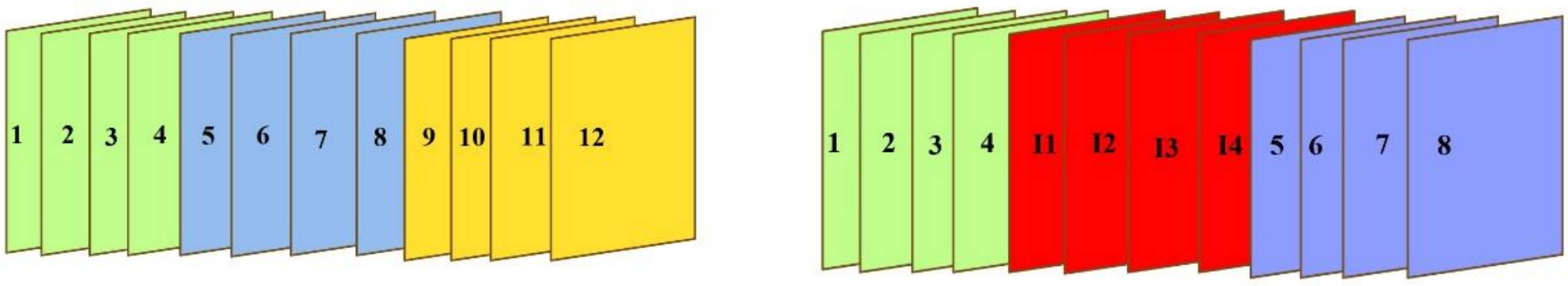

(**a**) (**b**)

**Figure 17.** Frame Insertion Forgery: (**a**) Original Video Sequence. (**b**) Frame Insertion Tampering.

Lastly, the frame shuffling or replication method, this counterfeit rearranges or changes the original arrangement of video frames, giving the original video a false meaning [88]. Figure 18 demonstrates the sequence of the forged video before and after involving the frame shuffling tampering.

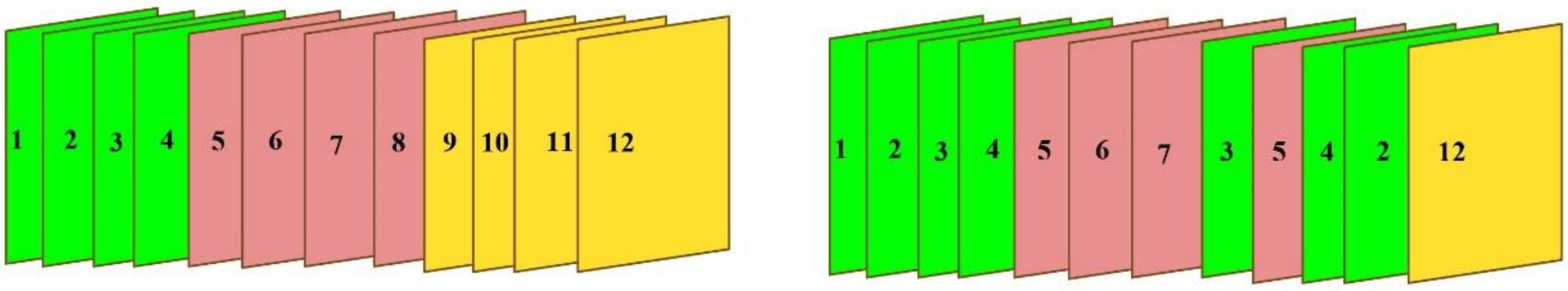

(**a**) (**b**)

**Figure 18.** Frame Shuffling Forgery: (**a**) Original Video Sequence. (**b**) Frame Shuffling Tampering.

2.2.3. Spatio-Temporal Tampering (Forgery)

Spatio-Temporal Tampering combines both methods that are explained above in 2.2.1 and 2.2.2 It simply modifies the combined sequence of frames and the information present in the same video frames [89]. Figure 19 demonstrates the procedure of applying spatiotemporal forgery to obtain fake videos.

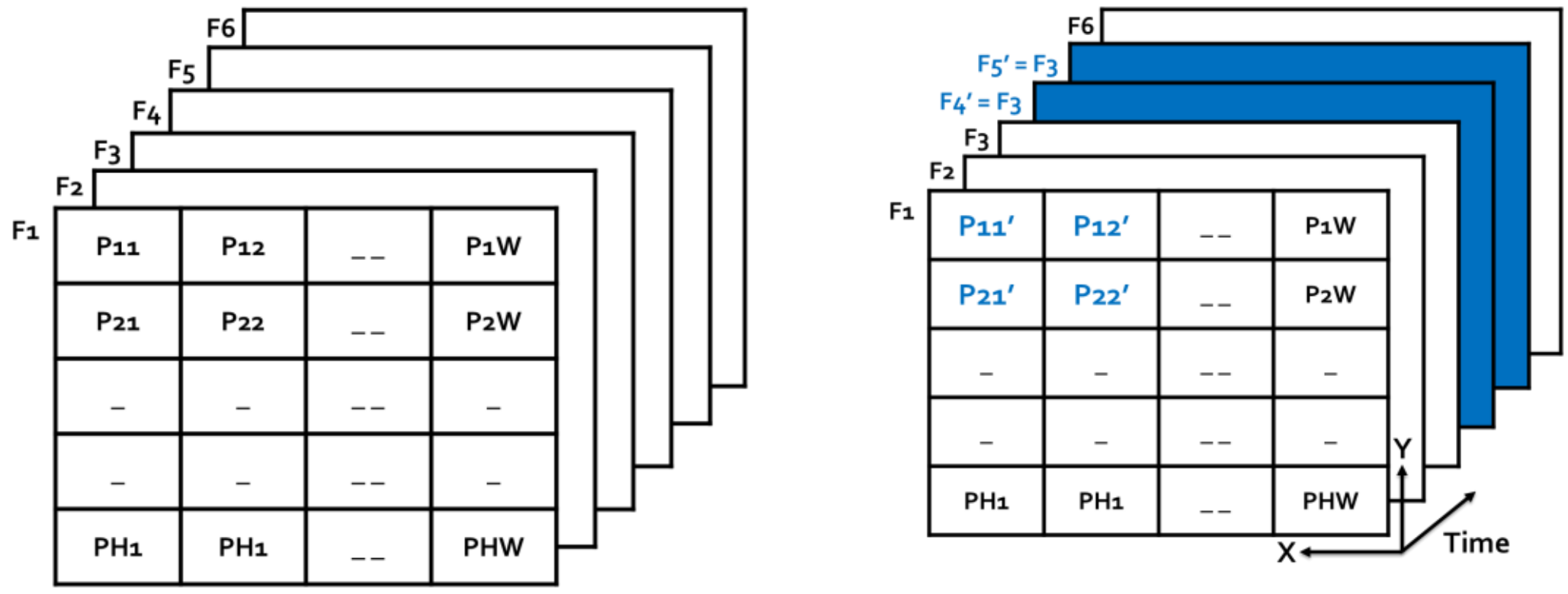

(a) (b)

**Figure 19.** Spatio-Temporal Video Tampering: (**a**) Original Video. (**b**) Spatio-Temporal Video Tampering.

2.2.4. Re-projection

Re-projection [90] is the act of recording a movie from the theatre screen to breach copyright laws [91]. Since the quality of the recorded video is deficient, it is undoubtedly effortless to detect. Figure 20 - a illustrates the place of the audience that recorded the film Live Free or Die Hard in the cinema Hall. While Figure 20 - b contains perspective distortion because of how the video camera is angled concerning the screen during the re-projected process. The camera skew, a distortion based on the angle between both the horizontal and vertical pixel axes, can be introduced into the intrinsic camera characteristics [92].

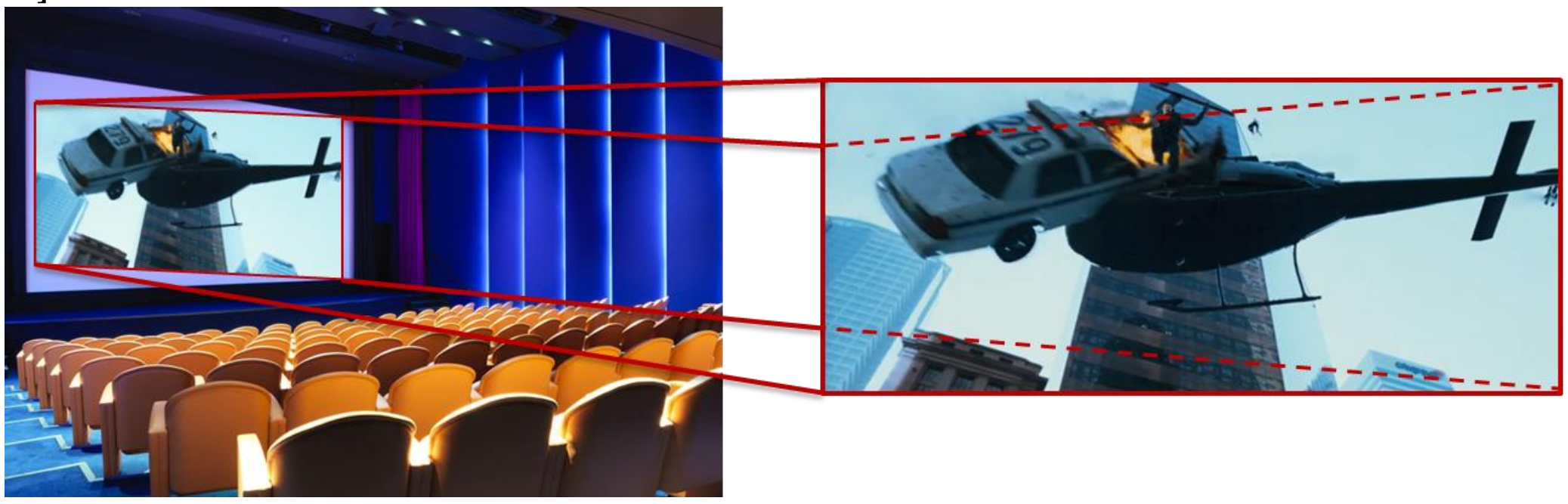

(a) (b)

**Figure 20.** Re-Projected Video: (**a**) Original scene from the movie Live Free or Die Hard presented in the cinema hall. (**b**) The projected movie that has been recorded. It illustrates the distortions which can be technically utilized to detect video tampering.

## 3. Forgery Detection in Surveillance Videos

This section surveys all the mechanisms that have been proposed to find tampering in videos that have been recorded using CCTV cameras. Table 3 delineates all the techniques and approaches that have been proposed so far to detect tampering in surveillance videos.

Approaches and techniques for forgery detection in surveillance videos are listed below:

1. Sensor Pattern Noise Technique
2. Gaussian Distribution
3. Residual Gradient and Optical Flow Gradient
4. Residual Frames
5. Optical Flow Gradient and Residual Analysis
6. Feature Extraction
7. WiFi Signals
8. Temporal Domain
9. Capsule Network
10. Secure-Pose

11. Similarity Analysis
12. Deep Learning
13. Radio-Frequency (RF) Signal

*3.1 Sensor Pattern Noise Technique*

Sensor Pattern Noise (SPN) [93] [122] and resampling estimation [94] techniques have been proposed to identify fakes in surveillance footage. Minimum Average Correlation Energy - Mellin Radial Harmonic (MACE-MRH) correlation filters can detect upscale crop, partial manipulation, and video alternation forgeries by utilising invariance or scaling tolerance. This approach is also used to identify the source camera. In the first stage, the source camera for a certain video is recognised. Then, in the second step, the scalar factor and correlation coefficient are used to identify tampering in the videos. Videos of static scenes have considerably remarkably outperformed others. This method produced significantly superior results compared to Chen's method [95] (i.e., 15% higher accuracy particularly when the scaling factor for infrared video is 1.8) [96]. The previously mentioned approach has been improved in which the scaling tolerance of a Minimum Average Correlation Energy - Mellin Radial Harmonic (MACE-MRH) correlation has been filtered to consistently reveal video upscale-crop fraud and recognise partially altered portions.

According to this, since resampling creates specific statistical correlations in the provided content, its presence can be determined by checking for these correlations. The Sensor Pattern Noise (SPN) [97] has been utilized as a forensic feature and examined the differences between reference SPN and SPN of upscaled frames in terms of correlation characteristics. The approach was evaluated on a total of 1920 fabricated sequences constructed from 120 self-recorded RGB and infrared H.264 encoded test videos. As long as the scale and quality parameters were regularly checked and adjusted, this method achieved a TNR (True Negative Rate) of 100% and a TPR (True Positive Rate) of greater than 98%. In the instance of partial modification detection, the detection accuracy of 100% for dynamic scene videos and 94.2 to 100% for static scene videos was recorded for region sizes between 100 and 150 square pixels. This technique was found to be reliable while dealing with compressed videos in addition to RGB and infrared videos. It works with movies of both dynamic and static scenes taken with moving and not moving cameras [98].

*3.2 Gaussian Distribution*

In the optical-flow-based forgery detection approach, the probability distributions of optical-flow variations for unaltered surveillance videos were modelled using a Gaussian distribution. An anomaly was any irregularity in the flow fluctuations, and a statistical inference test (Grubb's test) was used to assign an anomaly score to the optical flow patterns of each test video. The degree to which the pattern demonstrated anomalous behavior determined this score. Lastly, to detect inter-frame forgeries, three cut-off levels (one for frame insertion, one for frame deletion, and one for frame duplication) were applied to the anomaly score, which identified the abnormalities. The technique was assessed using a total of 160 test clips, all of which were produced from two original MPEG-2-encoded videos extracted from TRECVID's [99] surveillance event detection data set. The detection accuracies for frame deletion, insertion, and duplication were determined to be 75%, 85%, and 82.5%, respectively. The reported accuracy rates for forgery localization were 96.9%, 100%, and 86.2%, respectively [100].

*3.3 Residual Gradient and Optical Flow Gradient*

For H.264 and MPEG-2 encoded films, a detection technique for inter-frame forgeries employing prediction residual gradient and optical flow gradient has been given. A hybrid technique based on motion and brightness gradient characteristics is used to determine forgeries by identifying variations between nearby frames, notably for manually mobile recorded films and surveillance footage. Using the spike count regardless of the number of frames in the video, the proposed technique automatically detects video manipulation. This method achieved an accuracy of 83% [101].

*3.4 Residual Frames*

For the detection and localization of digital video inter-frame duplication, another approach based on the idea of residual frames has been developed. To detect and discover frame duplication frauds, the entropy of DCT coefficients in each residual frame's standard deviation value is computed, and the similarity between pairs of feature vectors is assessed. Using positive predictive value (PPV), true positive rate (TPR), and F1 Score, the efficacy of this method has been tested. This technique can effectively detect inter-frame duplication tampering in an extremely short time and it obtained the following results after using the SULFA dataset: PPV: 98%, TPR: 99%, F1: 98%, and after using the VIRAT dataset obtained PPV: 97%, TPR: 98%, F1: 97% [102] [123].

*3.5 Optical Flow Gradient and Residual Analysis*

The forgery detection technique based on optical flow gradient characteristics and prediction residual analysis has been described. The approach can detect and identify video frame deletion, insertion, and duplication. When the video is altered, the temporal correlations between neighboring frames are broken, which is evaluated by the researchers. The window-based paradigm is used to locate the counterfeit. The suggested method is optimized for H.264 video and MPEG-2 codecs, and it is 83% accurate for both slow- and fast-motion video [103].

*3.6 Feature Extraction*

Feature extraction and novel point localization technique have been proposed. During the phase of feature extraction, the 2-D phase congruency of each frame was identified as a desirable image property. The relationship between adjacent frames was then determined. In the second step, anomalous places were identified using a clustering technique (k-means). The normal and abnormal points were divided into two groups. The average accuracy acquired for the 1st dataset [104] is 97.08% and for the 2nd dataset [105] is 93.13% [106].

*3.7 WiFi Signals*

It has been demonstrated that Wi-Fi signals are beneficial for revealing video looping assaults on surveillance systems. It utilizes handcrafted event-level timing and frequency information from time-series Wi-Fi and camera data, resulting in a slow reaction time and an inability to do fine-grained false localization. Consequently, none of the existing solutions simultaneously meet the real-time and fine-grained criteria of forgery detection and localization in video surveillance systems. SurFi is used to analyses event-level timing information from Wi-Fi and camera data to detect camera looping attacks. SurFi utilizes existing Wi-Fi infrastructure (requiring no further hardware or deployment costs) to extract channel state information (CSI), which is then analyzed and linked with video and CSI signals to identify discrepancies. SurFi can identify assaults with up to 95.1% accuracy [107].

*3.8 Temporal Domain*

Another approach for detecting inter-frame forging (i.e., frame deletion, insertion, and shuffling) has been presented, in which the manipulation takes place in the temporal domain. This approach uses the universal image quality index (UQI) of temporal averages (TP) for non-overlapping neighboring frames to detect illegitimate actions in an exceptionally short amount of time. Individual frames will be collected from the security camera's directly captured footage. Then, each frame's TP will be measured. Due to the consistency and regularity of the video, the UQI of every two adjacent TP images is used to extract unusual activity as illegal candidates; if the video is subjected to deletion, insertion, or shuffling, the similarity will decrease, and the Q values at the border of the doctored clip will be lower than those of other clips. Lastly, the least Q value of the related frames of TP candidates and their neighbors is used to select the locations of inter-frame attacks. The detection performance of UQI method using Precision, Recall, and F1 Score using frame deletion is 0.98, 0.99, and 0.98 respectively. The performance of the same metrics under frame-insertion tampering is 0.99, 0.99, and 0.99 respectively. However, with frame-shuffling forgeries, the performance of the measures is 0.96, 0.97, and 0.96, respectively. The outcomes of each of the three assessment criteria were compared to the procedures in [108]–[111] and proved the best outcomes in terms of Precision, Recall, and F1 Score values. Moreover, the proposed technique has the shortest execution time compared to the previous techniques mentioned in [108]– [111] because it compares the temporal averages of non-overlapping subsequence frames rather than examining each frame individually [112].

Instead of all frames, the temporal average of each shot was employed to detect frame duplication. Grey-level co-occurrence matrix (GLCM) features were obtained for feature vectors, and the similarity between adjacent vectors was used to detect frame duplication. Despite the inclusion of post-processing activities with high false positives due to weak boundaries of duplicated frames, the suggested technique obtained an accuracy rate of 95% to 99% and a low running time. Without post-processing, the accuracy rates for frame duplication with shuffling (FDS) and frame duplication (FD) were 94% and 99%, respectively. This proposed technique has been evaluated on the SULFA [105] and EASIEST datasets [113].

A technique for identifying video tampering based on sensor pattern noise in video frames has been presented. Denoising video frames yielded the noise patterns, which were then averaged to identify sensor noise patterns. Using a locally adaptable DCT, the sensor noise patterns were analyzed (Discrete Cosine Transform). To detect if a video was genuine or faked, the correlation of noise residues from several video frames was calculated. The method was evaluated on a dataset containing noise patterns and yielded satisfactory results; nevertheless, these findings are contingent on the physical specifications of the source device. The accuracy of the prior model is 96.6 % [114].

### *3.9 Capsule Network*

Based on Capsule Networks, a new digital forensic method for identifying object-based counterfeiting in surveillance recordings has been developed. Intra-frame and inter-frame statistical features of the video sequence have been recovered as the input of the capsule network utilizing motion residual computed for each video frame. The experimental results demonstrate that the proposed method, with different bit rate values and dataset resolutions, achieves significant performance in terms of Video Detection Accuracy (VDA), Authentic Frame Detection Accuracy (AFDA), Forged Frame Detection Accuracy (FFDA), and Double-compressed Frame Detection Accuracy (DFDA), regardless of the group image length and degree of video compression. With a 3 M bit rate and 1280 720 dataset resolution, for example, VDA: 100%, AFDA: 99.30%, DFDA: 97.94%, and FFDA: 84.97%. For a 1.5 M bit rate and 1280 720 dataset resolution, the VDA accuracy is 99.99%, the AFDA accuracy is 98.64%, the DFDA accuracy is 96.12%, and the FFDA accuracy is 81.05%. Although the accuracies for 3 M bit rate and 640 × 360 dataset resolution are VDA: 100%, AFDA: 98.95%, DFDA: 97.49%, and FFDA: 84.56% [115]. The results mentioned for VDA, DFDA, and FFDA are considered the best compared to [116] and [117].

### *3.10 Secure-Pose*

Secure-Pose, the novel cross-modal system that identifies and localizes forgery attacks in each frame of live surveillance video, has been implemented. In a half-hour, they generated their dataset by gathering multimodal data. Faster-RCNN is utilized for intra-frame assaults to detect and clip out a human item before replacing it with the equivalent blank backdrop segment. Their test data covered a forgery detection accuracy of 95% [118].

### *3.11 Similarity Analysis*

The AIFDT-SV-BAS approach for identifying inter-frame manipulation is based on a study of similarity that is not affected by a single or many scenes. The recommended method involves examining the suspicious video for scene transitions. Whenever a situation changes, the method separates the scene into many shots. The images are then fed into a passive-blind technique based on a similarity analysis [73]. However, if there is no scene change, the splitting is incomplete. Primarily, the histogram difference between two consecutive frames in the HSV color space will be used to detect forgeries. In addition, H-S and S-V color histograms can identify various variations. The proposed technique AIFDT-SV-BAS has been assessed using CASIA 2 and NC 16 datasets. Furthermore, this technique has been evaluated using precision, recall, and accuracy metrics. This method has significantly outperformed the benchmark [73] result with a precision of 98.07%, a recall of 100%, and an accuracy of 99.1% due to the scene change recognition and video segmentation before checking for counterfeit [119].

*3.12 Deep Learning*

A system for identifying inter-frame forgeries that segments a movie into video shots and fuses spatial and temporal information to generate a single picture for each shot has been created. For effective extraction of spatiotemporal features, a pre-trained 2D-CNN model is utilized. The structural similarity index (SSIM) is then utilized to construct deep-learning video features. Lastly, they used 2D-CNN and RBF Multiclass Support Vector Machine (RBF-MSVM) to detect temporal manipulation in the video. To detect inter-frame forgery, a dataset of 13135 videos containing three types of forged videos under different conditions was created using original videos from the VRAT, SULFA, LASIESTA, and IVY datasets. The dataset achieved TPRs of 0.987, 0.999, and 0.985 for frame deletion, insertion, and duplication, respectively [120].

*3.13 Radiofrequency (RF) Signals*

Learning-based algorithms have been designed to detect video forgery attacks using radiofrequency (RF) signals. It is an extended version of Secure-Pose [118]. The secure-Pose method identifies camera looping attacks by analyzing event-level timing and frequency data derived from the coexistence of Wi-Fi signal and camera data. However, it cannot give fast identification and precise location of forgeries. Subsequently, the enhanced study employed the RF-based approach, which identifies anomalous items with a detection accuracy of 98.7% and correctly localizes them during playback and manipulation [121].

## 4. Discussion

Table 4 below summarizes all the methods and techniques that have been implemented to detect forgery in surveillance videos.

**Table 1.** Summary of Forgery Detection Methods in Surveillance Videos.

| Ref | Detection Methods | Forgery Detection Details | Accuracy | Limitations | Year |
|---|---|---|---|---|---|
| [96] | Sensor Pattern Noise (SPN) and resampling estimation method. | Minimum Average Correlation Energy - Mellin Radial Harmonic (MACE-MRH) correlation filters may identify upscale crops, partial manipulation, and video alteration frauds by utilizing scaling tolerance or invariance. | 15% is more accurate than Chen's technique [95]. | No approach can detect the source device without forecasting the scaling factor and enhancing the accuracy of partially changed region prediction by examining the video's features, such as RGB/IR, dynamic/static scene, and compressed video. | 2013 |
| [98] | Identifying two tampering kinds upscale-crop and partial manipulation using Sensor Pattern Noise (SPN) | Employing the scaling tolerance of a minimum average correlation energy Mellin radial harmonic (MACE-MRH) correlation filter to consistently reveal video upscale-crop fraud and recognized partially altered portions in dynamic and static-scene videos (recorded using static and moving cameras) by omitting high-frequency components and adaptively determining the size of the window block. | For Dynamic Scene videos: 100%<br>For Static Scene videos: 94.2% -100% | The techniques considerably rely on a vast number of content-dependent parameters and thresholds. | 2013 |
| [100] | A method based on optical flow and anomaly detection is | The defined method is predicated on the notion that | Elevator Scene Deletion: 93.8% | The optical flow algorithm needs improvement to increase its ability | 2014 |
| | created to authenticate videos and identify the inter-frame forgery process (i.e. frame deletion, insertion, and duplication). | forgery operations result in discontinuity locations in the optical flow variation sequence. The qualities of these criteria vary depending on the type of fraud. To distinguish discontinuity spots, the technique of anomaly detection is utilized. | Insertion: 100%<br>Duplication: 86.8%<br>Airport Passageway Scene<br>Deletion: 100%<br>Insertion: 100%<br>Duplication: 85.7% | to detect abnormality and it is only allowed for videos with static backgrounds.<br>Furthermore, the estimation method needs enhancement to make it more sensitive to counterfeiting and less associated with natural variations in original videos. | |
| [101] | Employing prediction residual gradient and optical flow gradient, the suggested forensic method detected inter-frame forgeries in H.264 and MPEG-2 encoded videos, particularly manually mobile-recorded videos and surveillance footage. | The mechanism was developed to identify faked videos automatically by merely counting the spikes. It was determined that this method is independent of the motion of objects in a video sequence, the number of frames altered, the number of objects in a video sequence, illumination change, recording equipment, and compression codec. | 83% | The performance of the mechanism degrades when videos with relatively slow motion are applied.<br>The technique efficiency for videos with quite low quality needs to be improved. | 2017 |
| [102] | Using the standard deviation of residual frames to detect Inter-frame duplication detection. | Calculating the entropy of discrete cosine transform (DCT) coefficients for every residual frame to illustrate their different characteristics will lead to identifying duplicated frames using subsequence feature analysis. | Using SULFA Dataset:<br>PPV: 98%, TPR: 99%, F1: 98%<br>Using VIRAT Dataset<br>PPV: 97%, TPR: 98%, F1: 97% | Has limitations in detecting other kinds of inter-frame forgery in the minimum time.<br>When the copied clip occurs in a static scene, this mechanism fails. | 2017 |
| [103] | The approach detects frame insertion, deletion, and duplication in MPEG-2 and H.264 encoded videos using prediction residual and optical flow inconsistencies. | The inter-frame forgery alters the temporal relationship between subsequent video frames. When these interruptions are properly examined utilizing prediction residual and optical flow, they could assist in identifying indicators of manufacture and pinpointing the particular location of the counterfeit in the presented video sequence. | Detection: 83%<br>Localization: 80% | When used on videos with intense lighting, performance suffers. | 2017 |
| [106] | Developing a system for detecting inter-frame forgeries based on 2-D Phase Congruency and K-Means Clustering. | Measuring the relationship between adjacent frames will result in the detection of discontinuous points produced by forged video using the K-Means | 1st Dataset: 97.08%<br>2nd Dataset: 93.13% | 1. The precision value of detecting frame deletion at the beginning and the end of the video is low.<br>2. Inserted frames cannot be recognised whether they were | 2018 |
| | | Clustering technique, which will classify normal and abnormal points. | | spliced from another video or copied from the same video. | |
| [107] | Defining SurFi using commonly available Wi-Fi signals to identify surveillance camera looping assaults in real-time. | SurFi utilizes existing Wi-Fi infrastructure (no new hardware or deployment costs are required) to extract human activities from channel state information (CSI), which is then processed and matched with video and CSI signals to identify mismatches. | 95.1% | The SurFi approach is unable to identify and pinpoint forgery assaults in individual video frames. | 2019 |
| [112] | It has been proposed to detect inter-frame forging (i.e., frame deletion, insertion, and shuffling) when the manipulation process occurs in the temporal domain. | This approach uses the universal image quality index (UQI) of temporal averages (TP) for non-overlapping neighboring frames and performs it in a very short amount of time to identify illegal actions. | Frame-Deletion Precision 0.98 Recall 0.99, F1 Score 0.98 Frame-Insertion Tampering Precision 0.99 Recall 0.99 F1 Score 0.99 Frame-Shuffling Forgery Precision 0.96 Recall 0.97 F1 Score 0.96 | The Precision, Recall, and F Score results degrade when modification in tampered videos is added such as noise. | 2019 |
| [113] | Identifying a robust technique for identifying inter-frame fraud based on a temporal average of each frame and a gray-level co-occurrence matrix (GLCM) constructed using statistical textural data. | For similarity matching, the correlation between identical feature vectors is calculated while taking image size into account. | Without post-processing, the suggested method's precision rate for frame duplication (FD) is 94%, and for frame duplication and shuffling (FDS) is 99%. However, the precision rate with post-processing is 95% for (FD) and 99% for (FDS). | Since inaccurate shots result in a rise in false positives, shot boundary detection is low. | 2020 |
| [114] | Expanding an approach for detecting video manipulation using Sensor Pattern Noise (SPN) in video frames. | Extracting the noise residue by subtracting denoised frames from the original frames. The sensor pattern noise will then be generated by averaging these noises. Finally, the correlation of noise residues from separate video frames is computed to assess whether a video is legitimate or fabricated. | 96.61% | Detect a single kind of forged document. Residue for moving background videos is not appropriate. | 2020 |
| [115] | A new digital forgery detection approach has been introduced based on Capsule | Motion residual is extracted from video sequences to improve the performance of | 3 M bit rate and 1280 × 720 dataset resolution | It can identify the forged frames in a video but not their precise location in the frame. | 2021 |
| | Networks for the identification of object-based counterfeiting in surveillance recordings. | the proposed capsule network for assessing videos for genuine, double-compressed, and faked frames. | VDA: 100%, AFDA: 99.30%, DFDA: 97.94%, FFDA: 84.97% 1.5 M bit rate and 1280 × 720 dataset resolution VDA: 99.99%, AFDA: 98.64%, DFDA: 96.12%, FFDA: 81.05%.<br>3 M bit rate and 640 × 360 dataset resolution VDA: 100%, AFDA: 98.95%, DFDA: 97.49%, FFDA: 84.56% | | |
| [118] | Presenting Secure-Pose the revolutionary cross-modal system that identifies and localizes forging traces on each suspicious live surveillance video frame using ambient wireless signals. | The Secure-Pose system successfully extracts human pose characteristics from the time-series camera and Wi-Fi data. It successfully identifies and localises forging traces in each frame under both inter-frame and intra-frame assaults. | 95.1% | It is unable to do fine-grained forgery localization. | 2021 |
| [119] | An inter-frame video forgery method AIFDT-SV-BAS for detecting tampering in surveillance videos has been proposed. | This approach examines the suspicious video for scene changes. If a scenario changes, the approach divides the scene into several shots. The shots are then loaded into a passive-blind approach based on similarity analysis [73]. However, the splitting is not completed if there is no scene change. The histogram difference between two adjacent frames in the HSV color space will be used to detect forgeries. In addition, H-S and S-V color histograms can detect numerous variations. The proposed technique AIFDT-SV-BAS has been assessed using CASIA 2 and NC 16 datasets. | 99.1% | This technique obtained a similar accuracy report to the benchmark during single shots video. | 2022 |
| [120] | Presenting an inter-frame forgery detection system based on a 2D convolution neural network (2D-CNN) in | The video is analyzed using 2D-CNN and Gaussian RBF Multiclass Support Vector Machine (RBF-MSVM) to | True Positive Rate (TPR) in Detection Insertion, Deletion, and Duplication Forgery is | This method can detect single inter-frame tampering in one video. | 2022 |
| | which a video is segmented into video shots and spatial and temporal information is fused to generate a single picture of each image. | detect a temporal counterfeit. | 99.9%, 98.7%, and 98.5% respectively. | | |
| [121] | Determining the extended Secure-Pose approach, which employs the extensive coexistence of surveillance and Wi-Fi infrastructures to fight video forgery assaults in the real-time and fine-grained form via RF technology. | The new approach successfully identifies and localizes forging traces in video streams by extracting standard human semantic information from the synchronized camera and Wi-Fi signals. | 98.7% | - | 2022 |

It can be proven that the extended Secure-Pose [121] is the best mechanism implemented as surveillance cameras and Wi-Fi signals can extract human semantic information and effectively detect and localize tampering activities on the video feeds. Besides, the assessment findings show that improved Secure-Pose has a high accuracy of 98.7% and correctly detects and finds anomalous items during playback and tampering assaults.

## 5. Conclusion and Future Work

Due to the rapid growth of digital multimedia technology, the significance of multimedia data, such as digital images/videos, is increasing rapidly in a variety of domains, with surveillance video footage serving as the primary form of evidence in a great number of court cases, including those involving highly sensitive information. With the widespread availability of inexpensive software for image/video manipulation, such as Priemer Rush, Quik, and LumaFusion, digital images/videos are now particularly susceptible to alteration/modification attacks. Video forgery refers to any deliberate modification of a digital video for fabrication. Digital video forgery detection is the approach used to assess whether digital video footage has been manipulated intentionally. Because no prior work has reviewed the methods and strategies used to detect a forgery in surveillance videos, the primary purpose of this article is to cover all prior works applied to detect tampering in surveillance videos and to state their opposing sides.

It could be concluded that the improved version of Secure-Pose [121] is considered to be the best technique with an accuracy of 98.7%, as it could successfully identify and localize forging traces in the video streams by extracting standard human semantic information from a synchronized surveillance camera and Wi-Fi signals in real-time and fine-grained detail. In the future, we will try to get an actual CCTV footage dataset, apply all the mentioned methods, test, and compare their outcomes and results with the figures obtained so far from the stated approaches.

**Author Contributions:** Author Contributions: Conceptualization, Noor B. Tayfor and Tarik A. Rashid; methodology, Tarik A. Rashid; software, Bryar A. Hassan; validation, Shko M. Qader, Jafar Majidpour and Haval M. Sidqi; formal analysis, Mohammed H. Abdalla; investigation, Zaher M. Yaseen; resources, Ahmed S. Shamsaldin; data curation, Noor B. Tayfor; writing, original draft preparation, Mahmood Yashar Hamza; writing, review and editing, Aram M. Ahmed, Abdulhady A. Abdullah and Janmenjoy Nayak; visualization, Abdulrahman Salih; supervision, Tarik A. Rashid and Hussein M. Ali; project administration, Tarik A. Rashid; funding acquisition, Hussein M. Ali. All authors have read and agreed to the published version of the manuscript.